%% file: main.tex
\documentclass[final]{cvpr}

\usepackage{times}
\usepackage{epsfig}
\usepackage{graphicx}
\usepackage{amsmath}
\usepackage{amssymb}
\usepackage[english]{babel}
\usepackage{blindtext}
\usepackage{booktabs}
\usepackage{caption}
\usepackage{subcaption} 
\usepackage{multirow}
\usepackage{array}
\usepackage{graphbox}
\usepackage{comment}
\usepackage{microtype}
\usepackage[pagebackref=true,breaklinks=true,colorlinks,bookmarks=false]{hyperref}
\usepackage{multicol}

\usepackage[resetlabels]{multibib}
\newcites{supp}{References}

\begin{document}

%%%%%%%%% TITLE
\title{Analysis and Mitigations of Reverse Engineering Attacks\\on Local Feature Descriptors}
\input{authors.tex}
\maketitle

%%%%%%%%% ABSTRACT
\input{text/00-abstract.tex}

%%%%%%%%% CONTENT
\input{text/01-introduction.tex}
\input{text/02-related-work.tex}
\input{text/03-threat-model.tex}
\input{text/04-methodology.tex}
\input{text/05-evaluation.tex}
\input{text/06-conclusion.tex}

{\small
\bibliographystyle{ieee_fullname}
\bibliography{references}
}

\input{text/00-supplemental}

\end{document}

%% file: authors.tex
% arXiv submission author block
\author{Deeksha Dangwal$^{\dagger\ddagger}$, Vincent T. Lee$^\ddagger$, Hyo Jin Kim$^\ddagger$, Tianwei Shen$^\ddagger$, Meghan Cowan$^\ddagger$, Rajvi Shah$^\ddagger$, \\
Caroline Trippel$^\S$, Brandon Reagen$^*$, Timothy Sherwood$^\dagger$, Vasileios Balntas$^\ddagger$, Armin Alaghi$^\ddagger$, Eddy Ilg$^\ddagger$ \\
$^\dagger$University of California, Santa Barbara\\ 
$^\S{}$ Stanford University \\
$^*$ New York University \\
$^\ddagger$Facebook Reality Labs Research \\
\{deeksha, sherwood\}@cs.ucsb.edu, trippel@stanford.edu, bjr5@nyu.edu, \\ \{vtlee, hyojinkim, tianweishen, meghancowan, rajvishah, vassileios, alaghi, eddyilg\}@fb.com
}

%% file: text/00-abstract.tex
\begin{abstract}

As autonomous driving and augmented reality evolve, a practical concern is data privacy. In particular, these applications rely on localization based on user images. The widely adopted technology uses local feature descriptors, which are derived from the images and it was long thought that they could not be reverted back. However, recent work has demonstrated that under certain conditions reverse engineering attacks are possible and allow an adversary to reconstruct RGB images. This poses a potential risk to user privacy.
We take this a step further and model potential adversaries using a privacy threat model. Subsequently, we show under controlled conditions a reverse engineering attack on sparse feature maps and analyze the vulnerability of popular descriptors including FREAK, SIFT and SOSNet. Finally, we evaluate potential mitigation techniques that select a subset of descriptors to carefully balance privacy reconstruction risk while preserving image matching accuracy; our results show that similar accuracy can be obtained when revealing less information. 

\end{abstract}

%% file: text/01-introduction.tex
\section{Introduction}
\label{sec:introduction}

\input{figures/advertisement}

Privacy and security of user data has quickly become an important concern and a design consideration when engineering autonomous driving and augmented reality systems.
In order to support machine perception stacks, these systems require always-on information capture. 
Most of these use-cases rely directly or indirectly on the data that originates from the user's device, i.e., RGB, inertial, depth, and other sensor values. 
Data assets are potentially rich in private information, but due to the compute power limitations on the device, they must be sent to a service provider to enable services such as localization, and virtual content.
As a result, there is understandable concern that any data assets shared with a cloud service provider, no matter how well-trusted, can potentially be abused~\cite{cachin2009trusting}.
To enable augmented reality in practice, beyond the application functionality, privacy-preserving techniques are thus an important consideration. 

In this work, we focus on localization as a fundamental component of augmented reality.
Localization relies on visual data assets to make a prediction of the location and pose of the user; in particular, most established algorithms rely on local feature descriptors. Since these descriptors contain only derived information, they were long thought to be secure. 

Unfortunately, recent literature shows that descriptors can be reverse engineered surprisingly well. We show an example in Figure~\ref{fig:teaser}.
In general, a reverse engineering attack is the process by which an artificial object is deconstructed to reveal its designs, architecture, code or to extract knowledge from the object~\cite{eilam_eldad}. 
For feature descriptors, a reverse engineering attack attempts to reconstruct the original RGB image that was used to derive the feature descriptors.
The fidelity to which the original RGB image can be reconstructed roughly correlates to the severity of the potential risk to privacy.
Prior work \cite{weinzaepfel2011reconstructing,d2013bits,dosovitskiy2016inverting,invsfm} has shown that feature descriptors are potentially susceptible to such an attack under a range of conditions and configurations.
However, there is limited work on quantitatively analyzing privacy implications as well as evaluating potential defenses against such reverse engineering attacks, which our work will explore.

To scope the problem, we first outline a privacy threat model~\cite{linddun} to contextualize the practicality and data assets available to a descriptor reverse-engineering attack.
Using these assets, we show potential reverse engineering attacks and quantify the information leakage to evaluate the privacy implications.
We then propose mitigation techniques inspired by some of the current best practices in privacy and security~\cite{wuyts2015linddun}.
In particular, we propose two mitigation techniques: (1) reducing the number of features shared and (2) selective suppression of features around potentially sensitive objects.
We show that these techniques can mitigate the potency of reverse engineering attacks on feature descriptors to improve protections on user data.
In summary, we make the following contributions: 
\begin{enumerate}
    \item We present a privacy threat model for a reverse engineering attack to narrow down the privacy-critical information and scope the setup for a practical attack.  
    
    \item We demonstrate a reverse engineering attack to reconstruct RGB images from sparse feature descriptors such as FREAK~\cite{alahi2012freak}, SIFT~\cite{sift1999lowe} and SOSNet~\cite{sosnet}, and quantitatively analyze the privacy implications. In contrast to previous work~\cite{invsfm,dosovitskiy2016inverting}, our approach does not take additional information such as sparse RGB, depth, orientation, or scale as input.
    \item We present two mitigation techniques to improve local feature descriptor privacy by reducing the number of keypoints shared for localization. We show that there is a trade-off between enhanced privacy (less fidelity of reconstruction) and the utility (localization accuracy). We also show which keypoints are shared matters for privacy.
\end{enumerate}

%% file: figures/advertisement.tex
\begin{figure}[t]
\centering
     \begin{subfigure}[t]{0.15\textwidth}
         \includegraphics[align=c, width=\textwidth]{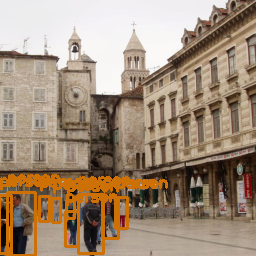}
         \vspace{-1mm}
         \caption{} 
         \vspace{-1mm}
     \end{subfigure}\hspace*{1mm}
     \begin{subfigure}[t]{0.15\textwidth}
         \includegraphics[align=c, width=\textwidth]{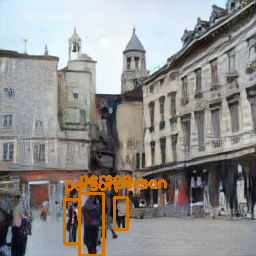}
         \vspace{-1mm}
         \caption{} 
         \vspace{-1mm}
     \end{subfigure}% 
    
    \vspace*{1mm}
    \begin{subfigure}[t]{0.15\textwidth}
         \includegraphics[align=c, width=\textwidth]{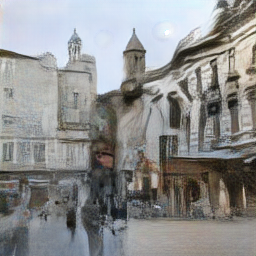}
         \vspace{-1mm}
         \caption{} 
         \vspace{-1mm}
     \end{subfigure}\hspace*{1mm}
     \begin{subfigure}[t]{0.15\textwidth}
         \includegraphics[align=c, width=\textwidth]{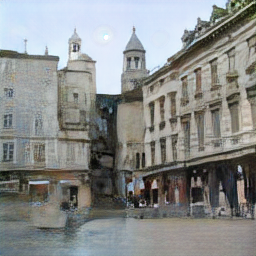}
         \vspace{-1mm}
         \caption{}
         \vspace{-1mm}
     \end{subfigure}% 
     \vspace{-2mm}
\caption{\textbf{Reverse Engineering Attack and Mitigations.} (a) Original image. Object detections are marked with orange bounding boxes.  (b) Image reconstruction from $1,000$ SIFT descriptors using our reverse engineering attack. The reconstruction gives sufficient fidelity to preserve detectable semantic information. By (c) reducing the number of features or (d) selective suppression around objects, we can reduce the efficacy of the attack and improve privacy.
\label{fig:teaser}
\vspace{-1.0em}
}
\end{figure}

%% file: text/02-related-work.tex
\section{Related Work}
\label{sec:prior-work}

The concept of reverse engineering local features has evolved over recent years as local descriptors play an increasingly important role.
Prior work focused primarily on better understanding the image features.
Only recently have there been proposals towards leveraging this line of research to understand the privacy implications.
Work towards discovering vulnerabilities and mitigating against attacks remains an emerging area of research.

\subsection{Recovering Images from Feature Vectors}

\noindent
\textbf{Reconstruction from Sparse Local Features}. 
Weinzaepfel et al.~\cite{weinzaepfel2011reconstructing} demonstrated the feasibility of reconstructing the input image, given SIFT~\cite{sift1999lowe} descriptors and their keypoint locations, by finding and stitching the nearest neighbors in a database of patches. 
d'Angelo et al.~\cite{d2013bits} cast the reconstruction problem as regularized deconvolution problem to recover the image content from binary descriptors, such as FREAK~\cite{alahi2012freak} and ORB~\cite{rublee2011orb}, and their keypoint locations. 
Kato and Harada~\cite{kato2014image} showed that it is possible to recover some of the structures of the original image from an aggregation of sparse local descriptors in bag-of-words (BoW) representation, even without keypoint locations.
While the quality of reconstructed images from the above methods is far from the original images, they allow clear interpretations of the semantic image content.
In this paper, we demonstrate that reverse engineering attacks using CNNs reveal much more image details and quantitatively analyse privacy implications for floating-point~\cite{sift1999lowe}, binary~\cite{alahi2012freak} and machine-learned descriptors~\cite{sosnet}.

\noindent
\textbf{Reconstruction from Dense Feature Maps}. 
Vondrick et al.~\cite{hoggles} perform a visualization of HoG~\cite{zhu2006fast} features in order to understand its gaps for recognition tasks.
To understand what information is captured in CNNs, Mahendran and Vedaldi~\cite{mahendran2015understanding} showed the inversions of CNN feature maps as well as a differentiable version of DenseSIFT~\cite{liu2010sift} and HoG~\cite{zhu2006fast} descriptors using gradient descent. 
Dosovitskiy and Brox~\cite{dosovitskiy2016inverting} took an alternative approach to directly model the inverse of feature extraction for HoG~\cite{zhu2006fast}, LBP~\cite{ojala2002multiresolution} and AlexNet~\cite{krizhevsky2017imagenet} using CNNs, and qualitatively show better reconstruction results than the gradient descent approach~\cite{mahendran2015understanding}. They also show reconstructions from SIFT~\cite{sift1999lowe} features using descriptor, keypoint, scale, and orientation information. All the above approaches differ from ours in that we perform the reconstruction from descriptors and keypoints only.  

\noindent
\textbf{Modern Reverse Engineering Attacks}.
In the context of 3D point clouds and the AR/VR applications built on top of them, a common formulation of the reverse engineering attack is to synthesize scene views given the 3D reconstruction information. Recent work by Pittaluga et al.~\cite{invsfm} showed that it is possible to reconstruct a scene from an arbitrary viewpoint from SfM models using the projected keypoints, sparse RGB values, depth, and descriptors. Our work extends this approach by considering only the modalities available to an attacker as input, which are keypoints and descriptors. 

\subsection{Defences and Mitigations}
\noindent
\textbf{Mitigations for Attacks on Sparse Local Features.}
For reverse engineering attacks on local features, one notable recent work \cite{linecloud, geppert2020privacy, shibuya2020privacy} proposes using line-based features to obfuscate the precise location of keypoints in the scene to make the reconstruction difficult.
The key idea is to lift every keypoint location to a line with a random direction, but passing through the original 2D \cite{geppert2020privacy} or 3D keypoints \cite{linecloud}. Since the feature location can be anywhere on a line, this alleviates privacy implications in the standard mapping and localization process.  
Shibuya et al.~\cite{shibuya2020privacy} later extended this approach for SLAM. Similarly, Dusmanu et al.~\cite{dusmanu2020privacy} represent a keypoint location as an affine subspace passing through the original point, as well as augmenting the subspace with adversarial feature samples, which makes it more difficult for an adversary to recover original image content.

\noindent
\textbf{Mitigations on Raw Images}. Apart from local features, other works try to alleviate the privacy concern around sharing raw images by perturbing the images~\cite{ren2018learning, Li_2019_CVPR_Workshops,butler2015privacy,ryoo2016privacy,raval2017protecting,wu2018towards,pittaluga2019learning,Wang_2019_CVPR_Workshops}. One way of achieving this is to mask out or replace the parts of images (e.g., faces) that may contain private information~\cite{vishwamitra2017blur,ren2018learning, Li_2019_CVPR_Workshops}. 
Another stream of work focuses on encoding schemes or degrading images to prevent recognition of private image content~\cite{butler2015privacy,ryoo2016privacy,raval2017protecting,wu2018towards,pittaluga2019learning,Wang_2019_CVPR_Workshops}.
A few cryptographic methods were proposed to encrypt visual content in a homomorphic way on local devices \cite{erkin2009privacy,sadeghi2009efficient,yonetani2017privacy}, which allows computing on encrypted data without decrypting. However, such methods are computationally expensive and it is not clear how to apply them to complex applications such as localization.

\subsection{Relationship to Adversarial Attacks on Neural Networks}

Recent work has shown that it is possible to trick deep learning models with adversarial inputs to induce incorrect outputs~\cite{szegedy2013intriguing, nguyen2015deep, biggio2013evasion, akhtar18}. 
For example, an adversarial attack may engineer a perceptually indistinguishable input image to trick a deep learning model into emitting an incorrect classification result.

Conceptually, these adversarial attacks are similar to the defense or mitigation strategies that we will propose, since state-of-the-art reverse engineering attacks on descriptors rely on deep learning models.
Our mitigation techniques modify inputs in a way to prevent the deep learning model used in the attack from accomplishing its objective --- reverse engineering the image.
However, unlike prior work in this space, our work lifts the insight that inputs can be modified to induce incorrect outputs and leverages it to \textbf{defend} against reverse engineering attacks instead of as an attack vector.

%% file: text/03-threat-model.tex
\section{System and Threat Definition}

In this section, we first define privacy and utility
as well as their trade-offs as they are discussed in this paper.
We also describe our privacy threat model, which defines
assumptions on adversary behavior and the conditions for a practical reverse engineering attack.

\subsection{Definitions}
\label{sec:threat-model}

\noindent\textbf{Privacy}. 
The LINDDUN privacy threat modeling methodology, one particular methodology in academic discussions, looks at privacy through the following properties~\cite{linddun}: linkability, identifiability, non-repudiation, detectability, information disclosure, content unawareness, and policy.
The idea behind LINDDUN is that whenever users share information, one or more of these privacy properties may be at risk.
That is relied on for the notion that minimizing the amount of shared
information improves privacy. 
However, precisely quantifying the impact on privacy is application-specific and can be implemented as a continuum, modulating the amount of
information to be shared as required.
In this work, references to privacy risk and/or threat applies specifically to reidentification risk that comes as a direct result of the reverse engineering attack; we describe and evaluate the trade-offs in Section~\ref{sec:measuring_utility_privacy})

\input{figures/threat_model_fig}

\noindent \textbf{Utility}.
Utility captures the accuracy (or performance) of an application or how useful a data asset is to an application.
Applications may have multiple utility functions to present a well-rounded understanding of the operation. 
Utility generally presents a trade-off with privacy considerations
as performance tends to increase with dataset size, e.g., ML training.
In our case, we use feature matching recall as a proxy for localization accuracy (see Section~\ref{sec:measuring_utility_privacy}).   

\noindent\textbf{Privacy-Utility Trade-Off}.
Applying privacy-preserving techniques can adversely affect utility. 
The ideal objective of the system is to have both high utility and higher privacy, but in practice there is a fundamental \textbf{trade-off} between the amount of information that one is willing to share and the utility one receives from sharing it. 
In our case, this means there is a trade-off between the desired localization accuracy (utility) and the images that may potentially be revealed (privacy).
The descriptor-based localization service offers a balance between privacy and utility; features sent to the server are still useful to the application pipeline but do not directly leak the rich information content of RGB images that may contain private information.

In certain cases where the definitions of utility and privacy are simple, this trade-off can be formalized and reasoned about analytically (e.g. $k$-anonymity~\cite{kanonoymity}). 
In larger systems this is not possible and we must \textbf{actively} play the roles of attacker and defender to model possible attacks and understand the potential risks to user privacy from reidentification. 
In computer security and privacy, this is the role of a \textit{privacy threat model}~\cite{salter1998toward, threatmodel, torr2005demystifying, ucedavelez2012real, moranawiley, saitta2005trike, linddun}. 

\subsection{Privacy Threat Model}
\label{sec:localization_privacy_threat_model}

Building a privacy threat model is application specific. 
For our localization use-case, the closest is the LINDDUN "hard privacy" threat model~\cite{linddun} where the objective is to share as little information as possible to a potential adversary.
At a high level, LINDDUN proposes building a dataflow diagram of a system, data assets, adversary, and potential attack vectors.
These are then used to audit potential threats that may impact privacy properties.
In our work, we focus on identifiability, detectability, and information disclosure, which are the most relevant to our reverse engineering attack on RGB images. % Do we need to scrub this?
\textit{Identifiability} refers to whether an adversary can identify items of interest.
\textit{Detectability} refers to whether an adversary can detect whether items exist or not.
\textit{Information disclosure} refers to whether information about the user is disclosed to an adversary who should not have access to it.
An adversary with an RGB image can observe information about each of these properties which poses a risk to privacy.

\noindent\textbf{System Definition and Sensitive Data Assets.}
Figure~\ref{fig:threat_model} shows the relevant components of our privacy threat model.
Our system follows a client-server architecture to process localization requests.
For localization, there are two primary data assets: (1) RGB images and (2) feature descriptors.
The client takes RGB images and derives feature descriptors which are shared with the server to query the user's location and pose from a map. 
We focus on protecting the RGB images as these are data assets which could be used to identify items of interest.
Descriptors are perceived as more private and more acceptable to share because they do not \textbf{directly} leak RGB information. 
However in Section~\ref{sec:reverse_engineering_attack_results}, we will show that indirectly this is not true.

\noindent\textbf{Adversary Definition and Potential Attacks.}
Our privacy threat model considers the service provider as an adversary (Figure~\ref{fig:threat_model}) that is \textit{honest-but-curious}, which is canonical in the security literature~\cite{gazelle}.
The honest-but-curious adversary is a legitimate participant in the system and executes the agreed upon application or service faithfully (as opposed to outright malicious behavior).
But, while fulfilling the service, the adversary is \textit{curious} and may use available data to learn information about the client. 
In our case, the adversary poses a risk to the client's privacy by reverse engineering the RGB images from feature descriptors.
This is possible because the adversary has access to similar data -- specifically feature descriptors and source RGB images -- and large scale compute resources.
Together, this means an adversary is capable of training deep-learning models (such as a reverse engineering model) to analyze data in a reasonable amount of time.

The goal of this paper is to understand how a client's protection against an honest-but-curious adversary capable of training deep learning models to reverse engineer RGB images from feature descriptors could be enhanced.

%% file: figures/threat_model_fig.tex
{
\setlength{\textfloatsep}{1mm}

\begin{figure*}[t]
\centering
\includegraphics[width=0.9\linewidth]{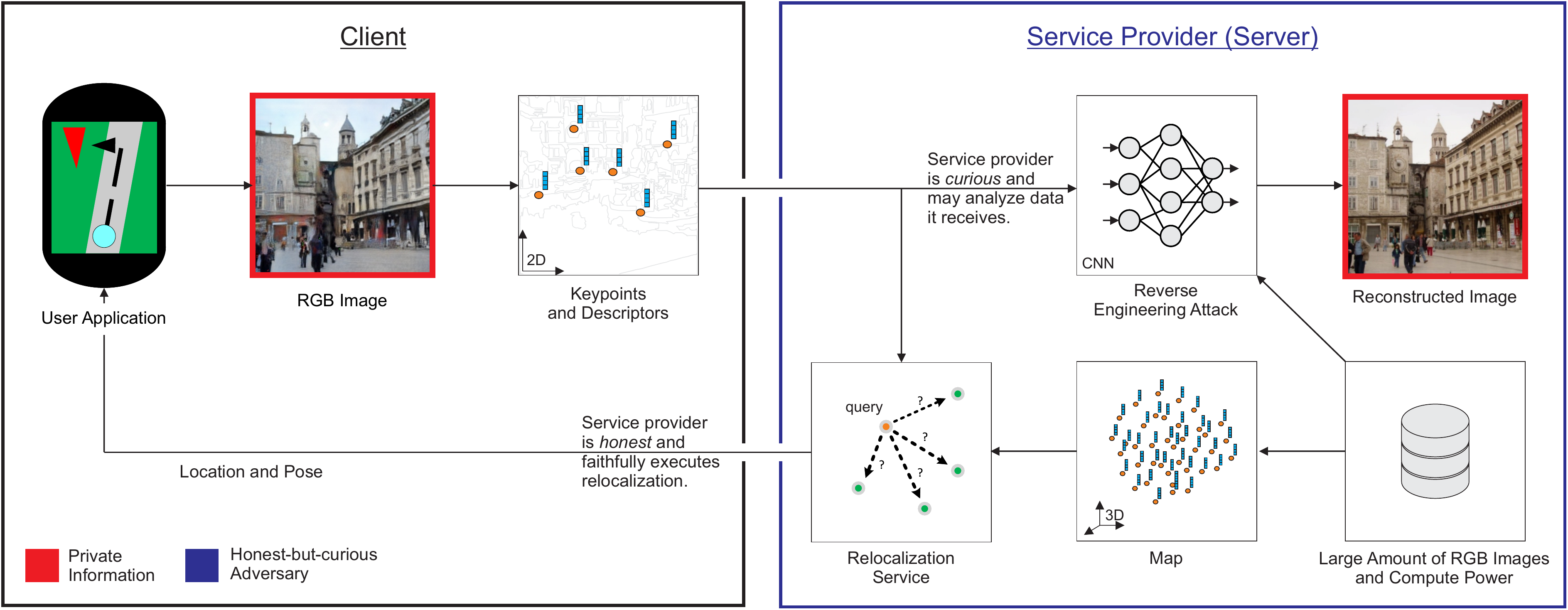}
     \vspace{-1mm}
\caption{\textbf{Overview of a typical localization system and our privacy threat model.} A client derives descriptors from RGB images and shares only the descriptors with a service provider. The service provider is honest and faithfully executes relocalization by matching query descriptors against those of a map. The service provider is also curious and may attempt to derive insight about the user. This sets up a potential reverse engineering attack. For such a system, our mitigation strategy must lie at the interface between the client device and the honest-but-curious service provider. Minimizing information shared at this interface reduces the amount of sensitive information that reaches the adversary.}

\label{fig:threat_model}
\vspace{-1.5em}

\end{figure*}
}

%% file: text/04-methodology.tex
\section{Reverse Engineering Attack}

\label{sec:methodology}

This section defines the convolutional neural network models we use to craft our reverse engineering attack.
As shown in Figure~\ref{fig:threat_model}, this model takes sparse local features (keypoints and descriptors) as input and estimates the original RGB image.

\subsection{Model Architecture}
\label{sec:architecture}

Given a user image $\mathbf{I}(i,j) \in \mathbb{R}^3$ and a derived sparse feature map $\mathbf{F}_{\mathbf{I},M}(i,j) \in \mathbb{R}^C$ containing $C$-dimensional local descriptors from the image $\textbf{I}$ using a feature extractor $M$, we seek to reconstruct an image $\mathbf{\hat{I}}(i,j) \in \mathbb{R}^3$ from $\mathbf{F}_{\mathbf{I},M}$. The sparse feature map is assembled by starting with zero vectors and placing extracted descriptors at keypoint locations $i, j$.
Our reverse engineering attack relies on a deep convolutional generator-discriminator architecture that is trained for each specific feature extraction method $M$.
The generator $G_M$ produces the reconstructed image: 
\begin{equation*}
    \mathbf{\hat{I}} = G_M(\mathbf{F}_{\mathbf{I},M})
\end{equation*}
and follows a single 2-dimensional U-Net topology~\cite{unet} with 5 encoding and 5 decoding layers as well as skip connections with convolutions. The discriminator $D_M$ is a 6 layer convolutional network operating on top of $G_M$~\cite{dcgann}. Please see the supplemental material for details.  In order to adhere to our privacy threat model and in contrast to prior work by Pittaluga et al.~\cite{invsfm}, we do not use depth or RGB inputs and subsequently also do not make use of a VisibNet. 

\subsection{Loss Functions}
We use the following loss functions to train the reconstruction network: 

\noindent \textbf{MAE.}
The mean absolute error (MAE) is the pixelwise L1 distance between the reconstructed and ground truth RGB images:
\begin{align}
    L_{mae} = \sum_{i,j} || \mathbf{\hat{I}}(i,j) - \mathbf{I}(i,j)||_1 \textrm{\,.}
\end{align}

\noindent \textbf{L2 Perceptual Loss.}
The L2 perceptual loss is measured as:
\begin{align}
    L_{perc} = \sum_{i,j} \sum_{k=1} ^ 3 ||\phi_k (\mathbf{\hat{I}}(i,j)) - \phi_k(\mathbf{I}(i,j)) ||_2^2 \textrm{\,,}
\end{align}
with $\phi_k$ being the outputs of a pre-trained and fixed VGG16 ImageNet model~\cite{imagenet}. $\phi_k$ are taken after the ReLU layer $k$ with $k\in\{2, 9, 16\}$. 

\noindent \textbf{BCE}.
For the generator-discriminator combination, we use the binary cross-entropy (BCE) loss defined as:
    \begin{align}
        L_{bce} = \sum_{i,j}  log(D_M(\mathbf{\hat{I}}(i,j))) + log(1 - D_M(\mathbf{I}(i,j))) \mathrm{\,.}
    \end{align}
Finally, we optimize the losses together:
\begin{align}
    L_{G} = L_{mae} +  \alpha L_{perc} + \beta L_{bce} \mathrm{\,,}
\end{align}
with $\alpha$ and $\beta$ as scaling factors.

\input{figures/reconstruction_figure}

%% file: figures/reconstruction_figure.tex
{
\setlength{\textfloatsep}{0mm}

\begin{figure*}[t]
    \centering
    \setlength{\tabcolsep}{0pt}

    \begin{tabular}{@{}lccccc@{}}
    
    \rotatebox{90}{\hspace{0.2cm} Original Image} &
    \includegraphics[width=0.15\linewidth]{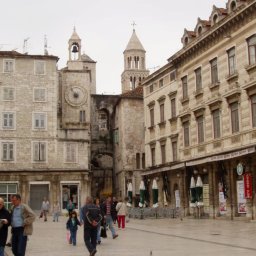} & 
    \includegraphics[width=0.15\linewidth]{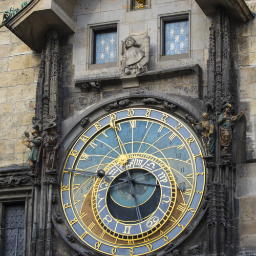} &
    \includegraphics[width=0.15\linewidth]{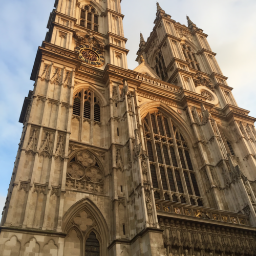} & 
    \includegraphics[width=0.15\linewidth]{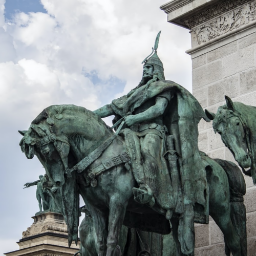} & 
    \includegraphics[width=0.15\linewidth]{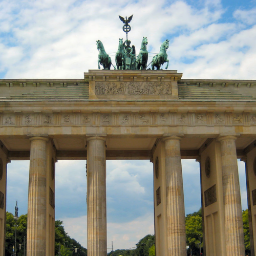} \\
    
    \rotatebox{90}{\hspace{0.8cm} SIFT} &
    \includegraphics[width=0.15\linewidth]{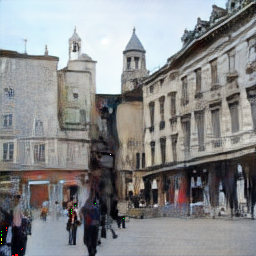} & 
    \includegraphics[width=0.15\linewidth]{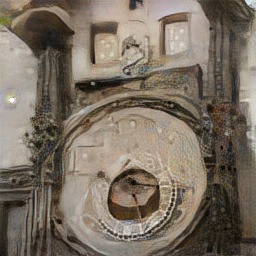} &
    \includegraphics[width=0.15\linewidth]{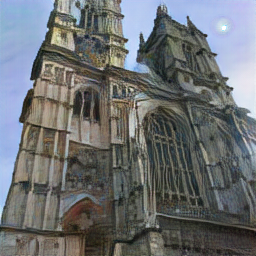} & 
    \includegraphics[width=0.15\linewidth]{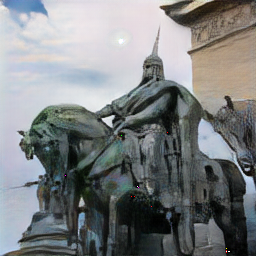} & 
    \includegraphics[width=0.15\linewidth]{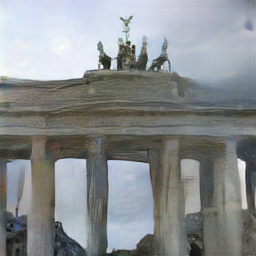} \\

    \rotatebox{90}{\hspace{0.6cm} FREAK} &
    \includegraphics[width=0.15\linewidth]{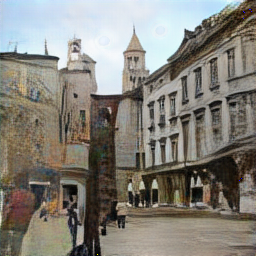} & 
    \includegraphics[width=0.15\linewidth]{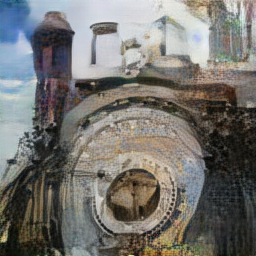} &
    \includegraphics[width=0.15\linewidth]{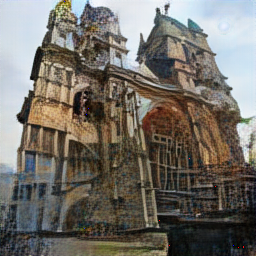} & 
    \includegraphics[width=0.15\linewidth]{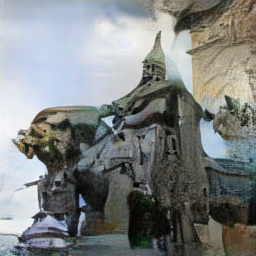} & 
    \includegraphics[width=0.15\linewidth]{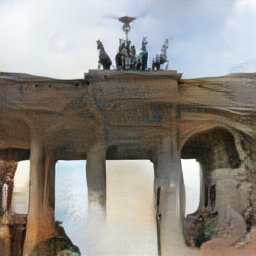} \\
    
    \rotatebox{90}{\hspace{0.6cm} SOSNet} &
    \includegraphics[width=0.15\linewidth]{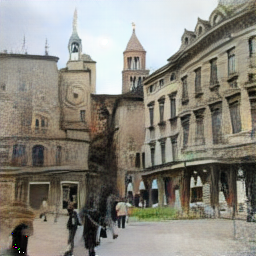} & 
    \includegraphics[width=0.15\linewidth]{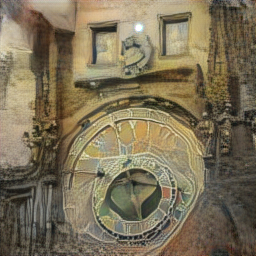} &
    \includegraphics[width=0.15\linewidth]{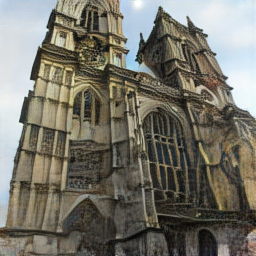} & 
    \includegraphics[width=0.15\linewidth]{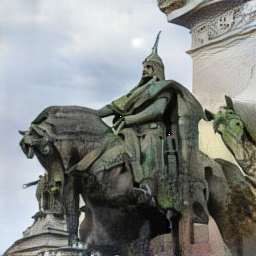} & 
    \includegraphics[width=0.15\linewidth]{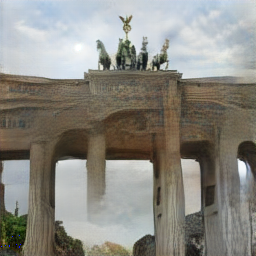} \\

    \end{tabular}
     \vspace{-4mm}
    \caption{\textbf{Reverse Engineering Attack Results.} From top to bottom: ground truth and reconstructions from a max. of $1,000$ sparse SIFT, FREAK and SOSNet features. One can observe that reconstruction from only sparse local features reveals the original image information extremely well. Note that the images show landmarks not included in the training data. Image attribution~\cite{jmiers_image}.}
    % Image attribution is required by license for middle image.
    \vspace{-1.5em}

    \label{fig:reverse_engineering_results}

\end{figure*}

}

%% file: text/05-evaluation.tex
\section{Evaluation}
\label{sec:evaluation}

\subsection{Experimental Setup}

\noindent
\textbf{Sparse Local Features.}
For the feature extraction method $M$ from Section~\ref{sec:architecture}, we use SIFT~\cite{sift1999lowe} ($C = 128$), FREAK~\cite{alahi2012freak} ($C = 64$), and SOSNet~\cite{sosnet} descriptors ($C = 128$) as representatives of traditional and machine-learned variants. 
Keypoint locations for FREAK and SOSNet were detected using Harris corner detection~\cite{harris_corner}.
For reconstruction, we use the SIFT detector for SIFT descriptors as in~\cite{invsfm}; however, for image matching we use Harris corners for SIFT descriptors because we found the SIFT detector performed poorly in this setting.

\noindent 
\textbf{Training and Evaluation Data.} 
We train our networks on $50,000$ images and their extracted sparse local features from the training partition of the MegaDepth dataset~\cite{megadepth}.
For testing the reverse engineering attack, we sampled $9,800$ images from the MegaDepth test set that contain objects as candidates for potential private data.

\noindent
\textbf{Network Training.}
A different reverse engineering model $M$ is trained for $400$ epochs for each descriptor type. 
The learning rate is initialized to $0.001$ and $0.0001$ for the generator and discriminator networks respectively.
Learning rates are adjusted using the Adam optimizer~\cite{adam_optimizer}.

\subsection{Measuring Privacy and Utility}
\label{sec:measuring_utility_privacy}

\noindent\textbf{Measuring Privacy with SSIM.}
Our first metric for measuring privacy is structural similarity (SSIM), which 
measures the perceptual similarity between images. In our case, we use SSIM to evaluate how much visual information the reverse engineering attack can recover by comparing against the original image. Therefore, SSIM provides a way to measure identifiability.  
We note that the SSIM measures to what extent the \textbf{whole} image may be recovered, which includes private and public information (e.g. people and buildings respectively); the public information is also available to the service provider when building the map. However, measuring how well the whole image can be reconstructed includes the reconstruction quality of private regions.
SSIM can further serve as a proxy to estimate how well other tasks such as object detection, landmark recognition, and optical character recognition may perform on the reverse-engineered image.

\noindent\textbf{Measuring Privacy by Object Detection.} 
We use an object detector (YOLO v3~\cite{redmon2018yolov3}, with 80 classes) to measure how much semantic information can be inferred from the reverse-engineered images. 
We compare object detection results on both the original and the reconstructed images.
If an object's bounding box in the original image has at least 50\% overlap with that of the reconstructed image of the same class label, we consider them as a match.
The more correspondence between objects in the original and the reconstructed image, 
the higher the risk to privacy.

\noindent
\textbf{Measuring Utility.}
To assess utility of local features when applying our mitigation strategies, we define an \emph{image matching} task as a proxy for localization and investigate how the feature matching between two images deteriorates as we increase the privacy. 
Specifically, we generate corresponding image pairs from the $53$ landmarks of the test split of the MegaDepth~\cite{megadepth} dataset. For each landmark, we sample $50$ pairs of images that have at least $20$ covisible 3D points determined from a reference map built with COLMAP~\cite{schoenberger2016sfm}, resulting in $2,650$ image pairs. For each corresponding pair of images, we perform local correspondence matching using input features, and count the number of pairs with at least $20$ inlier matches which we deem as successful. We refer to the proportion of image pairs that have been successfully matched as our matching recall, which we use as our utility measure.

\subsection{Reverse Engineering Attack}
\label{sec:reverse_engineering_attack_results}

\begin{table}[b]
\centering
\begin{tabular}{|l|c|c|} \hline
Descriptor & SSIM & Detected Objects \\
\hline \hline 
SIFT~\cite{sift1999lowe} & $0.675$ & $32.58\%$ \\
FREAK~\cite{alahi2012freak} & $\mathbf{0.511}$ & $\mathbf{19.32\%}$ \\
SOSNet~\cite{sosnet} & $0.616$ & $41.26\%$ \\
\hline
\end{tabular}
\caption{\textbf{Privacy metrics of reverse-engineered images using $1,000$ keypoints.} 
The amount of detected objects using YOLO v3~\cite{redmon2018yolov3} is measured on the reverse-engineered images relative the amount detected on the original images. 
FREAK descriptors reveal less information than SIFT and SOSNet.}
\label{tab:reverse_engineering_privacy_results}
\end{table}

\input{figures/keypoint_reduction}

We first evaluate to what extent the reverse-engineering attack from Section~\ref{sec:methodology} poses a reidentification risk to privacy.  
Examples of the reconstructions are shown in Figure~\ref{fig:reverse_engineering_results} and the privacy metrics of the reverse-engineered images are given in 
Table~\ref{tab:reverse_engineering_privacy_results}.
Reconstructions using FREAK~\cite{alahi2012freak} descriptors yield substantially poorer reconstruction quality and semantic content than SIFT~\cite{sift1999lowe} and SOSNet~\cite{sosnet}. %, thus FREAK can be considered more private.
Despite differences in feature extraction techniques and descriptor sizes, all three descriptors are susceptible to the attack and yield reconstructions comparable to prior work~\cite{invsfm} (please see supplemental material for detailed comparison to prior work), but notably without RGB or depth information as input.
At a higher level, the results show that under controlled conditions the reverse engineering attack can introduce a reidentification risk of RGB image content. 
The results from Table~\ref{tab:reverse_engineering_privacy_results} also show that the reverse-engineered images still allow an adversary to potentially detect and identify some objects that were present in the original images.

\subsection{Mitigation by Reduction of Features}
\label{sec:keypoint_reduction}

\begin{figure*}[t]
    \centering
    
    \begin{subfigure}[t]{0.33\textwidth}
        \centering
        \includegraphics[height=4.5cm]{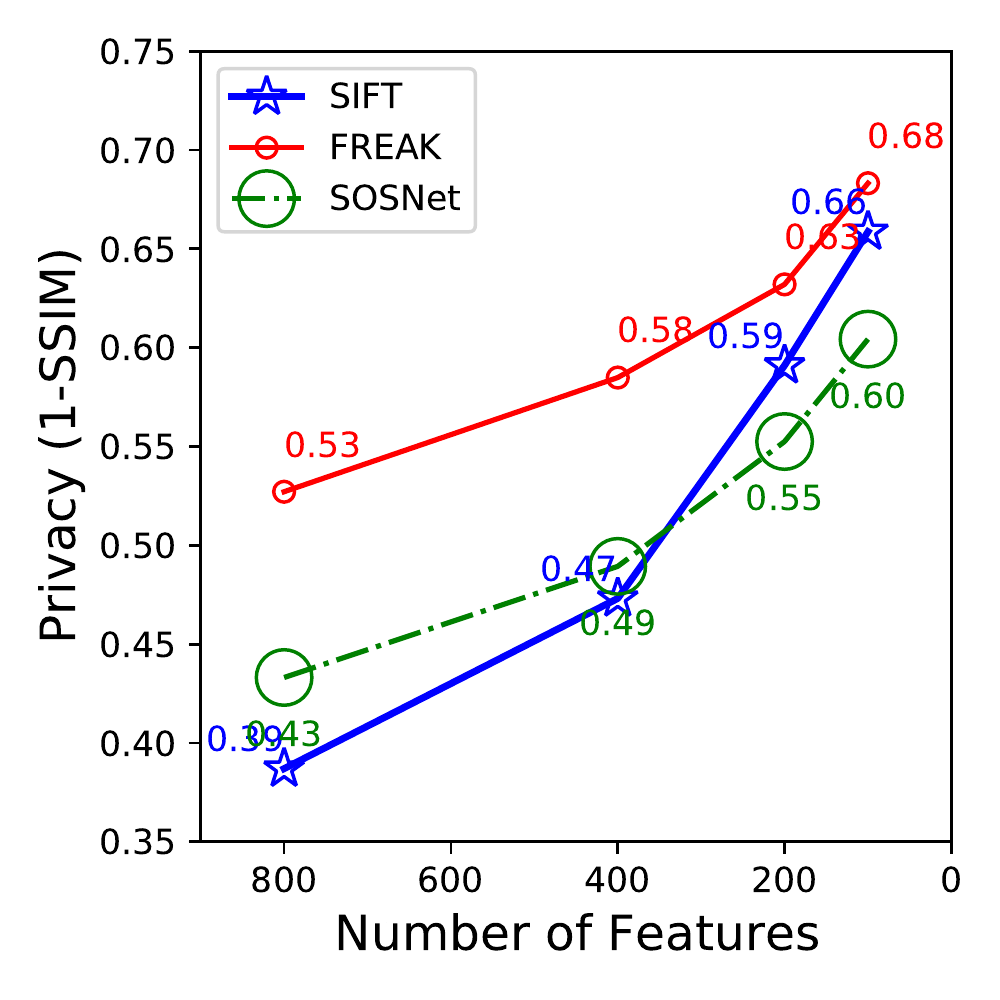}
        \vspace{-2mm}
        \caption{\label{fig:ssim_v_kp} 
        } 
    \end{subfigure}%
    \begin{subfigure}[t]{0.33\textwidth}
        \centering
        \includegraphics[height=4.5cm]{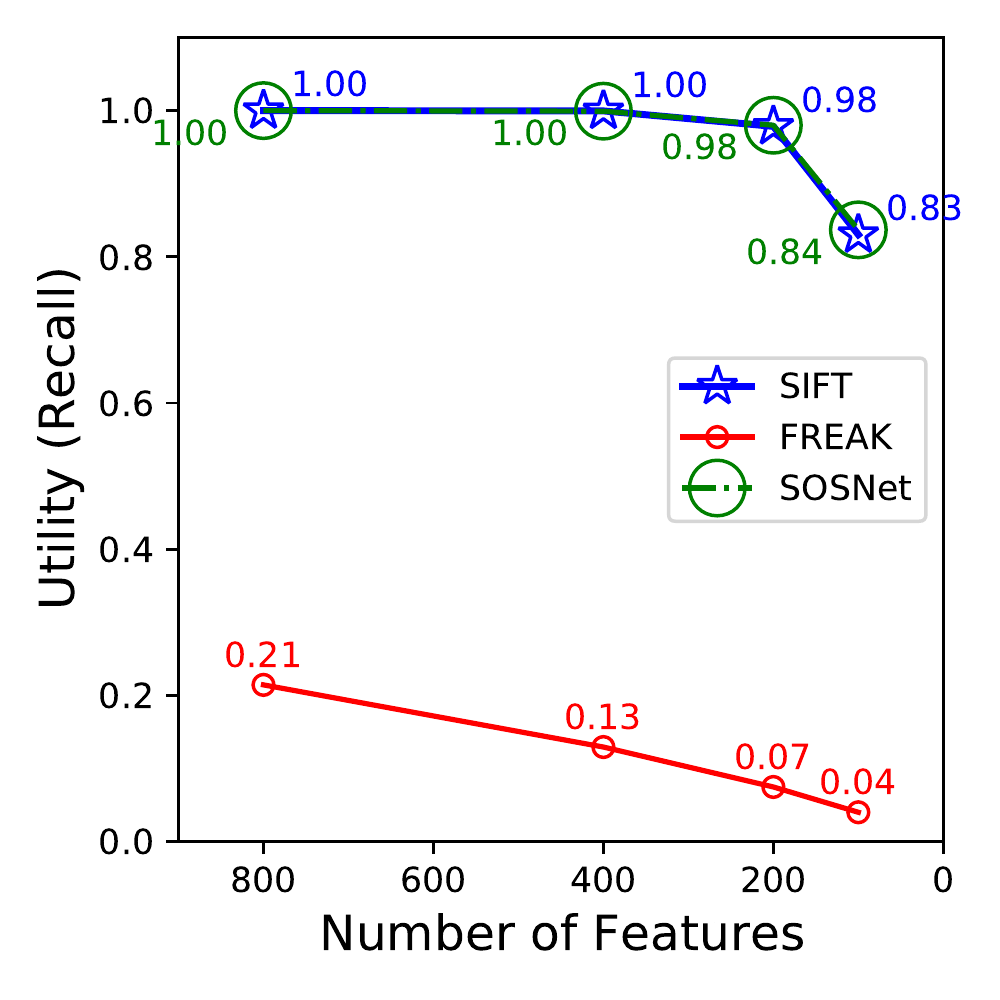}
        \vspace{-2mm}
        \caption{\label{fig:utility_v_kp}
        }
    \end{subfigure}%   
    \begin{subfigure}[t]{0.33\textwidth}
        \centering
        \includegraphics[height=4.5cm]{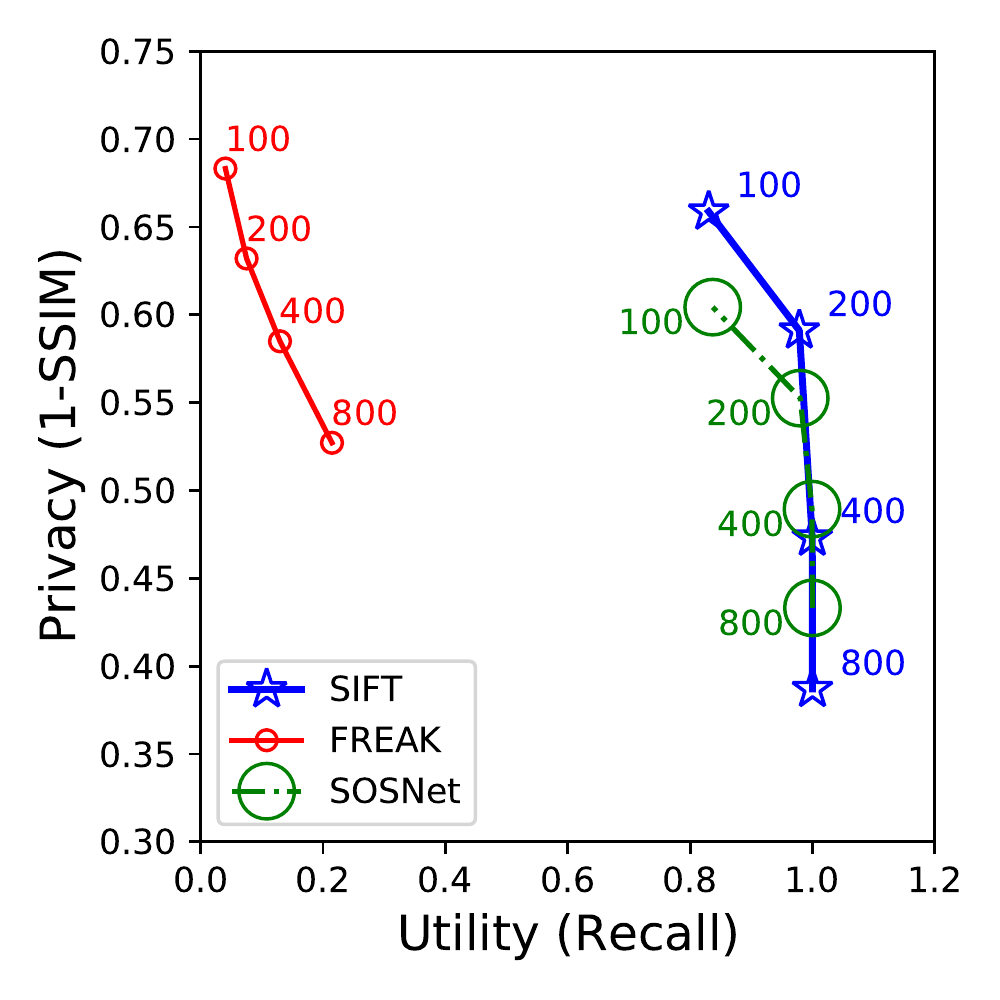}
        \vspace{-2mm}
        \caption{\label{fig:trade_off}
        }
    \end{subfigure}%
    \vspace{-3mm}
    \caption{
    \textbf{Utility and Privacy Trade-Off when Varying the Number of Features.}
    Privacy increases when reducing the number of features where FREAK gives the best results. For utility, FREAK and SIFT gives the best results. SIFT gives the best overall trade-off. 
    }
    \vspace{-5mm}

\end{figure*}

\input{figures/object_detection}

Following Section~\ref{sec:localization_privacy_threat_model}, to improve privacy, our objective is to minimize the information shared by the client. To this end, we investigate how reducing the number of features increases privacy at the expense of utility. 

For each descriptor type, we retain a maximum of $N$ top-scoring keypoints based on the detector response and vary $N$ from $1000$ to $100$. For each value of $N$ we then evaluate how well our reverse-engineering models perform. Qualitative results are given in~Figure~\ref{fig:keypoint_reduction}. 
We show the average privacy (measured by $1-$SSIM) of the reconstructed images vs. the number of features in~Figure~\ref{fig:ssim_v_kp}. The data shows the degradation in SSIM of the reconstructed images accelerates as more keypoints are removed. 
For less than $300$ features, SIFT gives better results than SOSNet. FREAK outperforms SIFT and SOSNet, and yields the best results in terms of privacy. 

However, despite strong privacy results, FREAK trades-off  utility.
In~Figure~\ref{fig:utility_v_kp}, we show how the utility changes. Here, FREAK gives the lowest utility, indicating that FREAK descriptors overall provide less useful information than SOSNet and SIFT. Interestingly, for SOSNet and SIFT the number of keypoints can be reduced to $200$ by sacrificing only $2\%$ performance. The trade-off between utility and privacy is shown in Figure~\ref{fig:trade_off}. Overall, we find that SIFT yields the best privacy-utility trade-off among the evaluated descriptor configurations on the Megadepth dataset.
We note that these results do not preclude the possibility that other descriptor configurations (i.e., in terms of dimensionality, target dataset, and type) may achieve better results.
Ultimately the ideal descriptor chosen will depend on the precise privacy and utility requirements necessitated by the localization service.

\subsection{Selective Suppression of Features}
\label{sec:keypoint_suppression}

\begin{table}[b]
    \centering 
    \begin{tabular}{|c||c|c||c|c|} 
    \hline 
     & \multicolumn{2}{c||}{Privacy} 
     & \multicolumn{2}{c|}{Utility}
     \\
     & \multicolumn{2}{c||}{(Object Recall)} 
     & \multicolumn{2}{c|}{(Matching Recall)}
     \\
    Supression & No & Yes & No & Yes \\ 
    \hline 
    \hline 
    SIFT~\cite{sift1999lowe}    & $20\%$ & $\mathbf{2.21}\%$ & $100\%$           & $\mathbf{88}\%$ \\
    FREAK~\cite{alahi2012freak}   & $11\%$ & $1.29\%$ & $\phantom{0}34\%$ & $28\%$ \\
    SOSNet~\cite{sosnet}  & $28\%$ & $5.21\%$ & $100\%$           & $88\%$ \\
    \hline 
    \end{tabular} 
 \caption{
        \textbf{Privacy-Utility Trade-Off for Selective Feature Suppression.}
        Object recall shows how many objects can be detected from the reverse engineered images compared to the original images without and with suppression (note that lower is better). 
        Matching recall shows how many images can be successfully matched without and with selective feature suppression. SIFT gives the best overall trade-off. 
        \label{tab:selective_trade_off}
    }   
   
\end{table}

Globally reducing image features can reduce the potency of the reconstruction attack, but at the same time it reduces the matching accuracy. In this section, we investigate to what extent an object detector can help implement a more selective approach. 
We identify and mark the sensitive regions in the images using the bounding boxes produced by the YOLO v3~\cite{redmon2018yolov3} object detector.
Based on the bounding boxes, we then suppress any features in these regions.
Finally, we apply our reverse-engineering attack and measure the detectable semantic information content in the images before and after reverse engineering (Table~\ref{tab:selective_trade_off}).

Figure~\ref{fig:object_detection} shows a qualitative example of how selective feature suppression effectively defeats the object detector; the people detected in the original image do not appear nor are identifiable by the object detector in the reconstructed images.
These results confirm our intuition that selective suppression can effectively preserve the privacy around a potentially sensitive region of interest (in our case semantic content of people in the image). Note that the quality of the overall image outside of the marked sensitive regions remains largely unaffected.
Finally, the results show that features of private objects should not be shared in order to mitigate privacy risks posed by reverse engineering attacks.

Results for the privacy-utility trade-off of the suppression are given in Table~\ref{tab:selective_trade_off}. Under the evaluated experimental conditions, SIFT and SOSNet give better trade-offs than FREAK; these trends are consistent with the results from Section~\ref{sec:keypoint_reduction}. 
Notably for SIFT the utility drops slightly, while the detected objects are almost eliminated.

%% file: figures/keypoint_reduction.tex
\begin{figure*}[t]
     \centering
{
    \setlength{\tabcolsep}{2pt}
    \begin{tabular}{@{}cccccccc@{}}
    
    & All Keypoints & 800 Keypoints & 400 Keypoints & 200 Keypoints & 100 Keypoints \\

    &
    \includegraphics[width=0.15\textwidth]{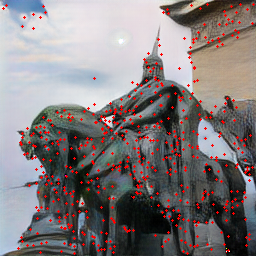} &
    \includegraphics[width=0.15\textwidth]{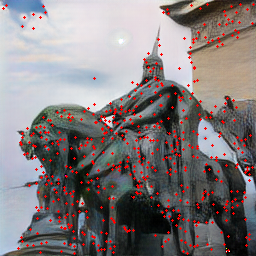} &    \includegraphics[width=0.15\textwidth]{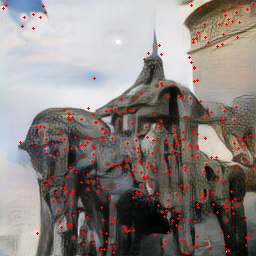} &    \includegraphics[width=0.15\textwidth]{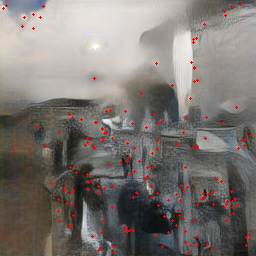} &    \includegraphics[width=0.15\textwidth]{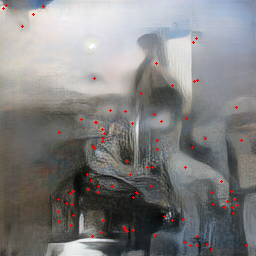} &
    \rotatebox{90}{\hspace{0.8cm} SIFT} \\
    Ground Truth & SSIM = 0.666 & SSIM = 0.666 & SSIM = 0.553 & SSIM = 0.375 & SSIM = 0.316 \\
     
    \includegraphics[width=0.15\textwidth]{figures/licensed_results/man_on_horse/gt.png} &

    \includegraphics[width=0.15\textwidth]{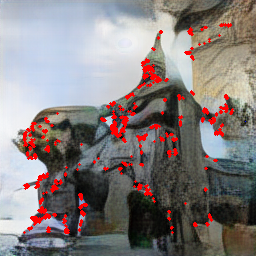} &
    \includegraphics[width=0.15\textwidth]{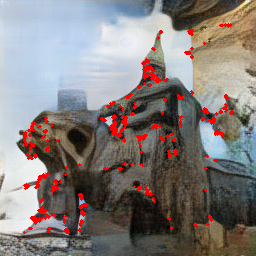} &    \includegraphics[width=0.15\textwidth]{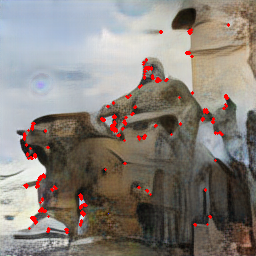} &    \includegraphics[width=0.15\textwidth]{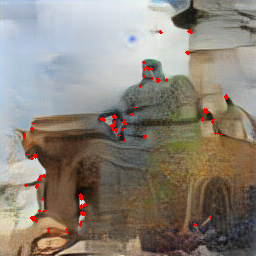} &    \includegraphics[width=0.15\textwidth]{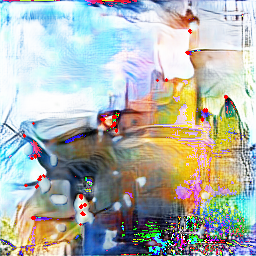} & 
    \rotatebox{90}{\hspace{0.6cm} FREAK}  \\  
    & SSIM = 0.488 & SSIM = 0.474 & SSIM = 0.406 & SSIM = 0.343 & SSIM = 0.300 \\
    
    &
    \includegraphics[width=0.15\textwidth]{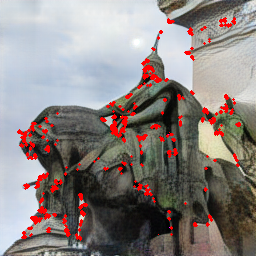} &
    \includegraphics[width=0.15\textwidth]{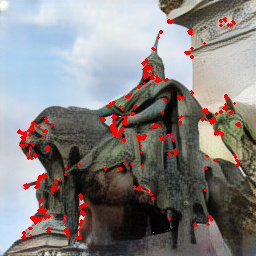} &    \includegraphics[width=0.15\textwidth]{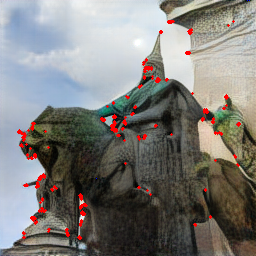} &    \includegraphics[width=0.15\textwidth]{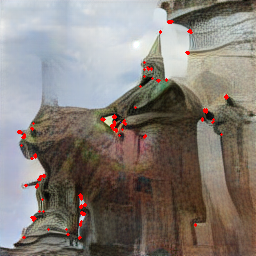} &    \includegraphics[width=0.15\textwidth]{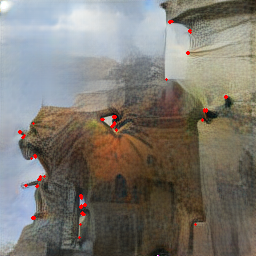} & 
    \rotatebox{90}{\hspace{0.6cm} SOSNet} \\
    & SSIM = 0.604 & SSIM = 0.582 & SSIM = 0.487 & SSIM = 0.407 & SSIM = 0.346 \\
    \end{tabular}
}
     \vspace{-3mm}
\caption{\textbf{Reverse engineering ablation study of reducing keypoints. } SIFT, FREAK and SOSNet reverse engineering results using $1,000$, $800$, $400$, $200$, and $100$ keypoints respectively, annotated in red. Reducing keypoints reduces the potency of the reverse engineering attack. Regions with higher densities of keypoints have better reconstruction quality.} 
\label{fig:keypoint_reduction}
    \vspace{-1.0em}

\end{figure*}

%% file: figures/object_detection.tex
\begin{figure*}[t]
    \setlength{\tabcolsep}{2pt}
    \centering
    \renewcommand{\arraystretch}{2}
    
    \begin{tabular}{@{} c c c c c @{}}
        \centering
        \multirow{10}{*}{
            \shortstack{(a) Original Image \\
            \includegraphics[width=0.17\linewidth]{figures/object_detection/ground_truth.png}}
        }
        & & SIFT & FREAK & SOSNet \\
     
        & (b) &
        \includegraphics[align=c, width=0.17\linewidth]{figures/object_detection/sift.png} &
        \includegraphics[align=c, width=0.17\textwidth]{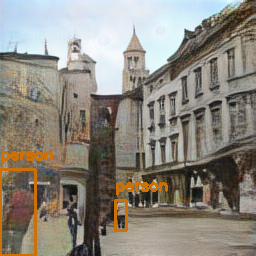} &
        \includegraphics[align=c, width=0.17\textwidth]{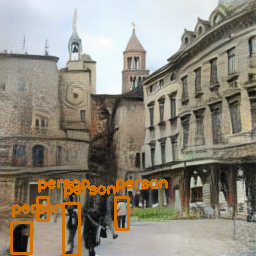} \\
        \vspace{-2.0em} \\
        
        \centering
        & (c) &
        \includegraphics[align=c, width=0.17\textwidth]{figures/object_detection/yhat_supp_sift.png} &
        \includegraphics[align=c, width=0.17\textwidth]{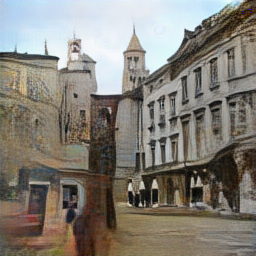} &
        \includegraphics[align=c, width=0.17\textwidth]{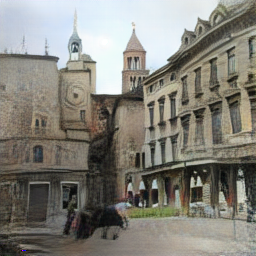} \\
     \end{tabular}
     \vspace{-2mm}
\caption{\textbf{Reverse Engineering Results after Selective Feature Supression.}
(a) Original image with object detection results (people). 
(b) Reverse-engineered images without feature suppression (with max. number of keypoints set to 1000), followed by object detection. 
(c) Reverse-engineered images using feature suppression, followed by object detection. 
As visible, all objects that can be detected by the YOLOv3 object detector without the suppression are successfully removed with the suppression.} 
\label{fig:object_detection}
    \vspace{-1.0em}

\end{figure*}

%% file: text/06-conclusion.tex
\section{Conclusion}
\label{sec:conclusion}

Our work has formulated a privacy threat model to scope the threats to descriptor-based localization.
In contrast to prior work, for the first time, we have shown a reverse engineering attack that operates in the real-world scenario, where only sparse local features are available to an honest-but-curious adversary. We found that our reverse engineering attack could reconstruct the original image with surprisingly good quality. We then investigated two mitigation techniques and showed a trade-off between privacy and utility (measured by feature matching). We found that using an object detector to suppress objects slightly reduces matching accuracy (as a proxy for localization accuracy) but gives better privacy results (fewer reidentifiable objects). Finally, our analysis has shown that, among the descriptors and we evaluate, the best overall privacy-utility trade-off can be achieved with SIFT, when compared to FREAK and SOSNet. 
Privacy (defined as reidentification risk through reverse engineering attacks as specifically described in this paper) may be preserved with the mitigation techniques described in this paper. 
Looking forward, our work provides initial experiments on some mitigation techniques the community may consider to further the privacy-aware descriptor-based applications research.

%% file: text/00-supplemental.tex
\setcounter{section}{0}
\renewcommand{\thesection}{\arabic{section}}

\begin{figure*}
\begin{center}
  \textbf{\Large Supplemental Material: Analysis and Mitigations of Reverse Engineering Attacks on Local Feature Descriptors}\\[0.8cm]
  
{ \large Deeksha Dangwal$^{\dagger\ddagger}$, Vincent T. Lee$^\ddagger$, Hyo Jin Kim$^\ddagger$, Tianwei Shen$^\ddagger$, Meghan Cowan$^\ddagger$, Rajvi Shah$^\ddagger$, \\
Caroline Trippel$^\S$, Brandon Reagen$^*$, Timothy Sherwood$^\dagger$, Vasileios Balntas$^\ddagger$, Armin Alaghi$^\ddagger$, Eddy Ilg$^\ddagger$\\[.1cm]
  
  { $^\dagger$University of California, Santa Barbara\\ 
$^\S{}$ Stanford University \\
$^*$ New York University \\
$^\ddagger$Facebook Reality Labs Research \\
\{deeksha, sherwood\}@cs.ucsb.edu, trippel@stanford.edu, bjr5@nyu.edu, \\ \{vtlee, hyojinkim, tianweishen, meghancowan, rajvishah, vassileios, alaghi, eddyilg\}@fb.com
}\\}
% (Dated: \today)\\[1cm]
\end{center}
\end{figure*}
\pagebreak
% \onecolumn
% Hi
% \vspace{-10cm}
% \twocolumn
% \begin{multicols}{2}

% \title{Supplemental Material: Analysis and Mitigations of Reverse Engineering \\ Attacks on Local Feature Descriptors}
% \maketitle

\section{Comparison to Prior Work}

\begin{figure*}[t]
\centering
     \begin{subfigure}[t]{0.23\textwidth}
         \includegraphics[align=c, width=\linewidth]{figures/related_work_comparison/ground_truth.png}
         \caption{Original} 
     \end{subfigure}\hspace*{1mm}
     \begin{subfigure}[t]{0.23\textwidth}
         \includegraphics[align=c, width=\textwidth]{figures/related_work_comparison/plaza_sift.png}
         \caption{Ours-SIFT\label{fig:ours_sift}} 
     \end{subfigure}\hspace*{1mm} 
    \begin{subfigure}[t]{0.23\textwidth}
         \includegraphics[align=c, width=\linewidth]{figures/related_work_comparison/plaza_freak.png}
         \caption{Ours-FREAK} 
     \end{subfigure}\hspace*{1mm}
     \begin{subfigure}[t]{0.23\textwidth}
         \includegraphics[ align=c, width=\textwidth]{figures/related_work_comparison/plaza_sosnet.png}
         \caption{Ours-SOSNet}
     \end{subfigure}% 

     \begin{subfigure}[t]{0.30\textwidth}
         \includegraphics[width=\textwidth]{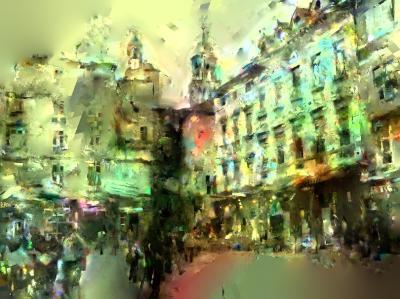}
         \caption{\protect\citesupp{weinzaepfel2011reconstructing}-SIFT\label{fig:weinz_sift}}
     \end{subfigure}\hspace*{1mm}%
     \begin{subfigure}[t]{0.30\textwidth}
         \includegraphics[width=\textwidth]{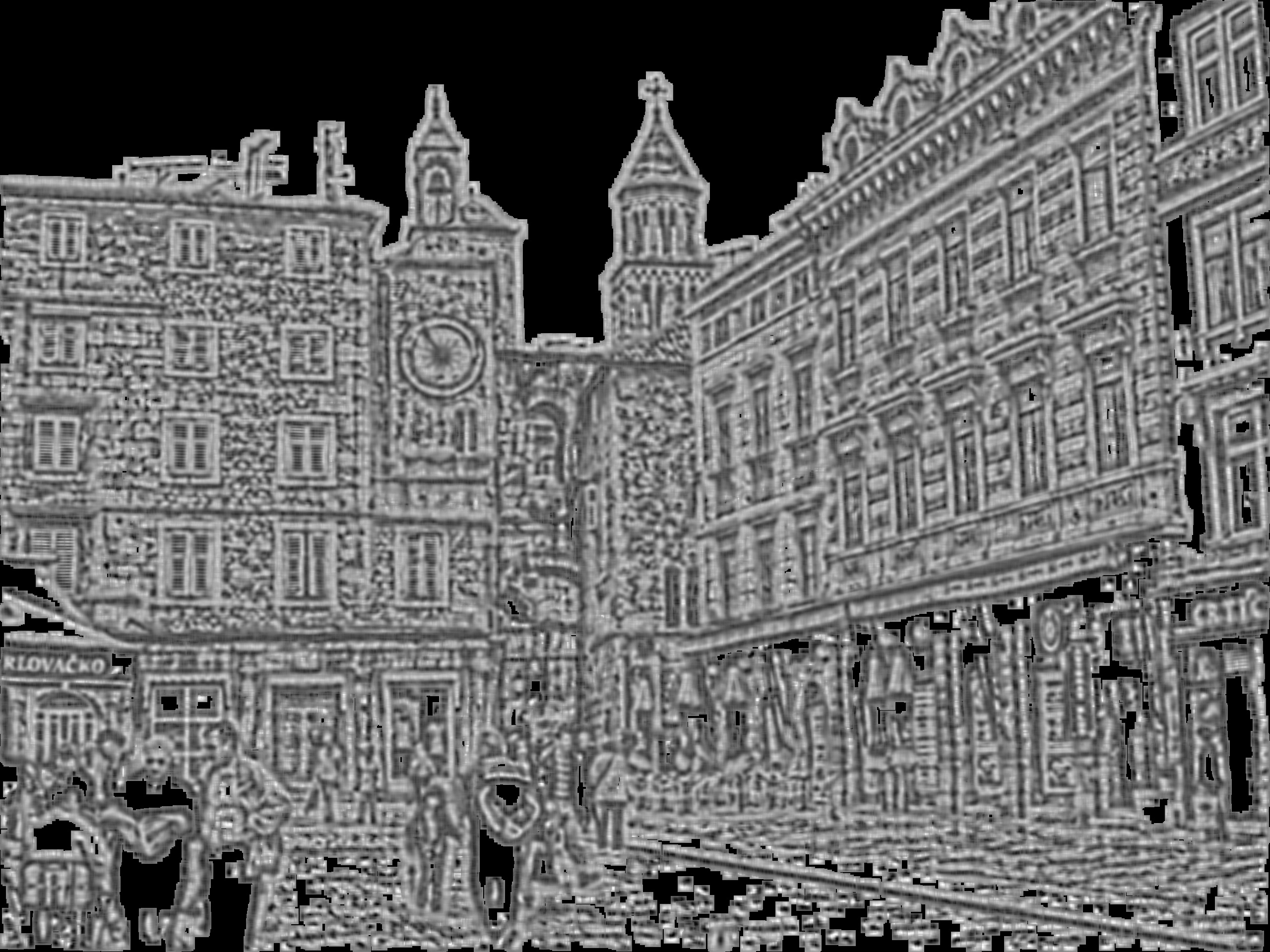}
         \caption{\protect\citesupp{d2013bits}-binary\label{fig:deangelo}}
     \end{subfigure}\hspace*{1mm}%
     \begin{subfigure}[t]{0.30\textwidth}
         \includegraphics[width=\textwidth]{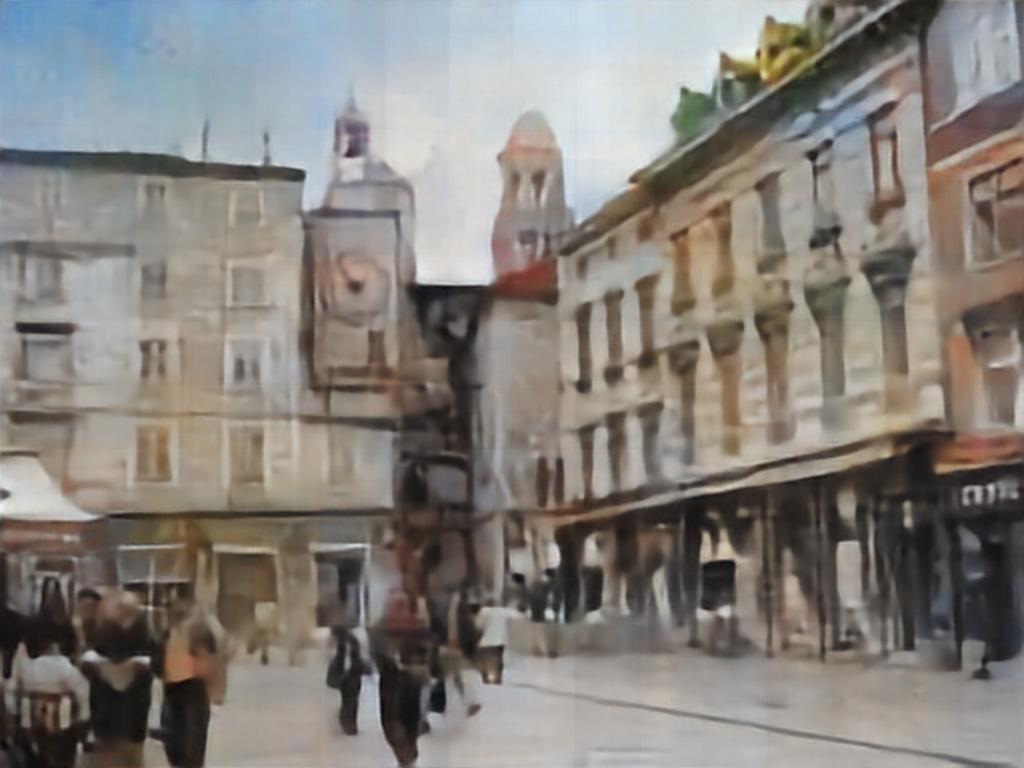}
         \caption{\protect\citesupp{dosovitskiy2016inverting}-SIFT\label{fig:dosovitskiy}}
     \end{subfigure} 
     \vspace{-2mm}
    \caption{(a) Original image and our reconstructions from SIFT (b), FREAK (c), and SOSNet (d) descriptors. Reconstructions by prior work from SIFT descriptors in~\protect\citesupp{weinzaepfel2011reconstructing} (e), and BRIEF descriptors in~\protect\citesupp{d2013bits} (f), and SIFT descriptors in~\protect\citesupp{dosovitskiy2016inverting} (g).
    }
    \label{fig:related_work_comparison}
\end{figure*}

We compare our work against several prior works that attempt to reverse engineer RGB images from features. 
\autoref{fig:related_work_comparison} compares our reverse-engineered image results compared to that of d'Angelo et al.~\citesupp{d2013bits} and Weinzaepfel et al. ~\citesupp{weinzaepfel2011reconstructing}.
Compared to the latter in~\autoref{fig:weinz_sift}, our result using SIFT shown in \autoref{fig:ours_sift} produces a qualitatively better reverse-engineered image with more accurate color estimates. As shown in~\autoref{fig:deangelo}, the work from d'Angelo et al. reconstructs image gradients only and is not comparable to our work. 
We also compare our results to those by Dosovitskiy and Brox~\citesupp{dosovitskiy2016inverting} in \autoref{fig:dosovitskiy}.
In contrast to our work, Dosovitskiy and Brox use more keypoints and descriptors for their reconstruction using SIFT descriptors; they use roughly $3000$ keypoints to reconstruct this image while we use $1,000$ or fewer in our experiments.
Qualitatively the results are comparable.

The previous state of the art is recent work proposed by Pittaluga et al.~\citesupp{pittaluga2019revealing} which also uses convolutional neural networks to reverse-engineer images. 
Pittaluga et al. use additional information such as depth and RGB at the keypoint location to supplement SIFT descriptors as input to their reverse engineering model.
Our work does not use depth nor RGB information, and does not make use of a separate network for visibility estimation (as the VisibNet from~\citesupp{pittaluga2019revealing}).
We also compare against FREAK and SOSNet descriptors while Pittaluga et al. exclusively analyze SIFT descriptors.

The results show that even without the additional depth and RGB information from Pittaluga et al., our reconstructions produce more detail and more accurate color in average in the cases of SIFT and SOSNet. In contrast, FREAK does not allow us to reconstruct the color information as well and we see some color artifacts (e.g., see the clock image).
Since a practical reverse engineering attack for a relocalization service does not provide depth or RGB information to the honest-but-curious adversary, our attack formulation aligns with the real-world scenario.
When using all input data assets (depth, SIFT, and RGB) Pittaluga et al. achieve a maximum average SSIM of $0.631$ on reconstructions and an average SSIM of $0.578$ when using only SIFT descriptors (\autoref{tab:ssim_results}).
In contrast, our reverse engineering attack yields an average SSIM of $0.675$ for reconstructions from SIFT features alone and thus provides a new state of the art. We attribute the improvements to our architecture choice and training procedure which we describe below.  

% \end{multicols}

\begin{table}[hbt!]
\centering
\begin{tabular}{@{}|l||l|c|@{}} \hline
& Inputs & SSIM \\ \hline \hline
\multirow{3}{*}{Prior Work~\protect\citesupp{invsfm}} & Depth Only & $0.578$ \\
& Depth+SIFT & $0.597$\\
& Depth+SIFT+RGB & $0.631$\\
\hline
\multirow{3}{*}{Ours} & SIFT Only & $0.675$\\
& FREAK Only & $0.511$ \\
& SOSNet Only & $0.616$ \\
\hline
\end{tabular}
\caption{Comparison of average SSIM values of the reverse engineered images from prior work~\protect\citesupp{invsfm} and our work. Our work achieves better SSIM results for SIFT without using inputs like depth or RGB.}
\label{tab:ssim_results}
\end{table}

\section{Architecture Implementation Details}
Our reverse engineering attack uses a deep convolutional generator-discriminator network (see main paper).  
We provide the implementation details of our reverse engineering network, including architecture, optimization, and training methodology in this section. 

\subsection{Generator} The generator follows a 2-dimensional U-Net~\citesupp{unet} topology with 5 encoding and 5 decoding layers. Specifically, the architecture of the encoder is $\mathrm{conv}_{64}$-$\mathrm{conv}_{128}$-$\mathrm{conv}_{256}$-$\mathrm{conv}_{512}$-$\mathrm{conv}_{1024}$, where $\mathrm{conv}_N$ denotes a convolutional layer with $N$ kernels of size $3 \times 3$, stride of 1, and padding of 1. A bias is added to the output, followed by a BatchNorm-2D, and ReLU operation. Between convolutions, there is a 2D MaxPool operation with kernel size and stride both set to 2. The decoder architecture is $\mathrm{upconv}_{1024}$-$\mathrm{upconv}_{512}$-$\mathrm{upconv}_{256}$-$\mathrm{upconv}_{128}$-$\mathrm{upconv}_{64}$ where $\mathrm{upconv}_N$ denotes a convolutional layer with $N$ kernels which is also upsampled by a scale factor of 2. The kernels for these layers are also $3 \times 3$ in size and have a stride and padding of both 1. The convolution is also followed by a BatchNorm-2D and ReLU operation.

\subsection{Discriminator} The discriminator used for adversarial training has the following architecture: $\mathrm{Disc}_{256}$-$\mathrm{Disc}_{128}$-$\mathrm{Disc}_{64}$-$\mathrm{Disc}_{32}$-$\mathrm{Disc}_{16}$-$\mathrm{Disc}_{8}$-$\mathrm{Disc}_{4}$ where $\mathrm{Disc}_N$ denotes a 2D-convolution with $N$ kernels of size $4 \times 4$, stride of 2, and padding of 1, followed by BatchNorm-2D and leaky ReLU with negative slope of $0.2$. $\mathrm{Disc}_{256}$ is not followed by a batch normalization and in $\mathrm{Disc}_{4}$ leaky ReLU is replaced by a sigmoid operation. 

\subsection{Training Methodology and Optimization}
The loss functions we use are described in Section 4.2 of our paper. Our losses together are described as:
\begin{align}
    L_{G} = L_{mae} +  \alpha L_{perc} + \beta L_{bce} \mathrm{\,,}
\end{align}
where, $\alpha = 1$, and $\beta = 0.1$. 

We detail how we use the L2 perceptual loss here. We utilize a VGG16 model pre-trained on ImageNet~\citesupp{deng2009imagenet}. The outputs of three ReLU layers are used: layers 2, 9, and 16. $\phi_i$ is used to denote the these layers. $\phi_1: \mathbb{R}^{H \times W \times 3} \to \mathbb{R}^{H/2 \times W/2 \times 64}$, $\phi_2: \mathbb{R}^{H/2 \times W/2 \times 64} \to \mathbb{R}^{H/4 \times W/4 \times 128}$, and $\phi_3: \mathbb{R}^{H/4 \times W/4 \times 128} \to \mathbb{R}^{H/8 \times W/8 \times 256}$. These outputs are used by the L2 perceptual loss to train the network.

Both the generator and discriminator were trained using the Adam optimizer with $\beta_1 = 0.9$ and $\beta_2 = 0.999$ and $\epsilon =1e^{-8}$. The learning rate for the generator is $0.001$ and for the discriminator is $0.0001$. We train each of the SIFT, FREAK, and SOSNet networks for $400$ epochs each. The first $250$ epochs are run without the discriminator contributing to the generator-discriminator combination network. The next $150$ epochs are run with both the generator and discriminator losses.  

{\small
\bibliographystylesupp{ieee_fullname}
\bibliographysupp{refs_supp}
}